\documentclass[11pt]{article}
\pdfoutput=1  

\usepackage[preprint]{acl}

\usepackage{times}
\usepackage{latexsym}

\usepackage[T1]{fontenc}

\usepackage[utf8]{inputenc}

\usepackage{microtype}

\usepackage{inconsolata}

\usepackage{graphicx}

\usepackage{amsmath}
\usepackage{amssymb}
\usepackage{color}
\usepackage{xspace}
\usepackage{multirow}
\usepackage{array}
\usepackage{makecell}
\usepackage{tabularx}
\usepackage{booktabs}
\usepackage[capitalize]{cleveref}
\providecommand{\yes}{\checkmark}
\providecommand{\no}{--}
\newcommand{\eg}{\textit{e.g.}\xspace}

\usepackage{placeins}

\makeatletter
\newcommand{\publiconly}[1]{\ifacl@anonymize\else#1\fi}

\makeatletter
\long\def\@makefntext#1{\noindent\@makefnmark\,#1}
\makeatother
\makeatother

\title{Telco-GAIA: Bilingual Benchmark for Agents in Telecom Domain}

\author{
  \textbf{Dmitrii Khizbullin\textsuperscript{1,*}},
  \textbf{Zaid Alyafeai\textsuperscript{1}},
  \textbf{Abdelrahman Eldesokey\textsuperscript{1}},
  \textbf{Nourah AlSultan\textsuperscript{2}},
\\
  \textbf{Raghad Alshalan\textsuperscript{2}},
  \textbf{Bernard Ghanem\textsuperscript{1}},
  \textbf{David R. Pugh\textsuperscript{1}}
\\
\\
  \textsuperscript{1}King Abdullah University of Science and Technology (KAUST),
  \textsuperscript{2}stc
\\
  \small{\textsuperscript{*}Corresponding author: \href{mailto:dmitrii.khizbullin@kaust.edu.sa}{dmitrii.khizbullin@kaust.edu.sa}}
}

\begin{document}
\maketitle
\begin{abstract}
We introduce Telco-GAIA, a bilingual, multi-modal benchmark for evaluating tool-using agents on the data of a real-world telecommunications operator. Telco-GAIA comprises 100 human-verified question-answering tasks, in English and Arabic, that each demand multi-hop reasoning (4.2 hops on average) over three heterogeneous sources: a static website snapshot (HTML, images, and linked PDFs), a synthetic relational SQL database, and external web archives, spanning text, image, and tabular modalities. The benchmark is delivered as a sandboxed Docker environment and scored by normalized exact string matching, making evaluation objective, deterministic, and reproducible over time without any LLM-as-a-Judge. Evaluating a purpose-built reference agent across twelve commercial and open LLMs, we find Telco-GAIA challenging: even the strongest model solves only 71\% of tasks; under a moderate cost budget, this falls to about 40\%, and the visually grounded categories remain the weakest, where the average backend scores below 30\%, leaving substantial headroom in document and image understanding. Telco-GAIA offers a rigorous, reproducible testbed for enterprise agents and a template for constructing closed-domain benchmarks.
\end{abstract}

\section{Introduction}

Large language model (LLM) agents are increasingly deployed in customer-facing enterprise settings, where they must navigate proprietary websites, parse internal documents, and query operational databases to resolve customer requests.
Telecommunications is a prototypical example of such a setting: a single operator exposes hundreds of plans, devices, roaming options, and promotions on a large bilingual website, backed by relational billing and usage data.
Determining whether an agent can reliably retrieve and reason over these heterogeneous sources is, therefore, a practical prerequisite for deployment.

Rigorous evaluation of such agents, however, remains difficult.
A common practice generates question-answer pairs semi-automatically from text chunks and scores responses with an LLM-as-a-Judge (LLMaJ).
This is convenient but neither objective nor reproducible: LLMaJ scores are sensitive to the judge model and to prompt wording~\cite{zheng2023judging,wang2024fair}. Moreover, automatically generated questions are not guaranteed to have a unique, verifiable answer: over half of open-domain questions admit multiple valid answers~\cite{min2020ambigqa}, and audits find 13--17\% of responses scored incorrect to be actually correct, penalised by gold-standard reference answers that are themselves incomplete or wrong~\cite{min2021efficientqa}.
Academic benchmarks address parts of the problem but leave a gap for this setting.
Open-domain agent benchmarks such as GAIA~\cite{mialon2023gaia} rely on the live internet and are thus not reproducible over time;
sandboxed web environments such as WebArena~\cite{zhou2024webarena} are general-purpose and do not combine relational-database, PDF, and image retrieval within a single domain;
and enterprise RAG benchmarks such as WixQA~\cite{cohen2025wixqa} are text-only, monolingual, and provide no live agent environment or structured data.
None targets a bilingual, multi-modal, telecom-specific corpus with deterministic scoring.

We introduce \textbf{Telco-GAIA}\publiconly{\footnote{\raggedright Questions and resources:\\
{\fontsize{7.4pt}{9pt}\selectfont\href{https://huggingface.co/datasets/kaust-generative-ai/telco-gaia}{\nolinkurl{hf.co/datasets/kaust-generative-ai/telco-gaia}}}\\
Ground-truth:\\
{\fontsize{7.4pt}{9pt}\selectfont\href{https://huggingface.co/datasets/kaust-generative-ai/telco-gaia-groundtruth}{\nolinkurl{hf.co/datasets/kaust-generative-ai/telco-gaia-groundtruth}}}\\
Leaderboard:\\
{\fontsize{7.4pt}{9pt}\selectfont\href{https://kaust-generative-ai-telco-gaia-leaderboard.static.hf.space/index.html}{\nolinkurl{telco-gaia-leaderboard.static.hf.space}}}}}, a benchmark for evaluating tool-using agents on the public data of a real telecommunications operator serving tens of millions of subscribers.
Telco-GAIA draws design inspiration from GAIA~\cite{mialon2023gaia} but targets a closed enterprise domain in a sandboxed, reproducible environment.
It comprises 100~multi-hop tasks (mean 4.2~hops) spanning seven categories (Pricing, Images, Web Archives, PDF, PDF Visual, Database, and Miscellaneous) in English and Arabic (65 and 35~tasks, respectively).
Each task has a single, human-verified correct answer scored by normalized exact string matching, removing any dependence on an LLM judge.
Solving a task requires retrieving and reasoning over three heterogeneous sources: a static snapshot of the operator's website (HTML pages, images, and linked PDFs), a synthetic relational SQLite customer database exposed via a REST~API, and external web archives (Wikipedia and ArXiv).
The website and database are served locally via Docker, while the web archives are virtually immutable; the environment is therefore semi-closed, and runs are reproducible across time. Because the operator's site is real and public, a model might answer some tasks from parametric memory rather than retrieving from the served snapshot; this shortcut only weakens over time, as the live site drifts from our frozen snapshot, forcing genuine retrieval and hardening the benchmark rather than decaying it; a no-tools ablation confirms the residual leakage is small (\Cref{sec:pk_mode}). \Cref{sec:harness_appendix} details the harness, evaluator, and architecture.

To establish baseline difficulty, we evaluate a purpose-built tool-calling reference agent across twelve commercial and open models.
The benchmark proves challenging: the strongest model reaches 71\% overall, and accuracy is lowest on the visually grounded categories (Images and PDF Visual), exposing a concrete weakness of current agents in document and image understanding.
\Cref{tab:benchmark_comparison} situates Telco-GAIA against prior benchmarks.

\begin{table*}[t]
\centering
\small
\begin{tabularx}{\textwidth}{Xccccc}
\toprule
\textbf{Benchmark} & \makecell{\textbf{Multi-}\\\textbf{modal}} & \makecell{\textbf{Telecom}\\\textbf{domain}} & \makecell{\textbf{Relational}\\\textbf{DB}} & \makecell{\textbf{Golden steps}\\\textbf{annotation}} & \makecell{\textbf{Multi-/bi-}\\\textbf{lingual}} \\
\midrule
GAIA~\cite{mialon2023gaia}               & \yes & \no  & \no  & \yes   & \no  \\
WebArena~\cite{zhou2024webarena}          & \no  & \no  & \no  & \no   & \no  \\
VisualWebArena~\cite{koh2024visualwebarena} & \yes & \no & \no & \no   & \no  \\
TheAgentCompany~\cite{xu2025theagentcompany} & \no & \no & \no & \yes$^\ddagger$ & \no  \\
WorkArena~\cite{drouin2024workarena}      & \no  & \no  & \no  & \no   & \no  \\
AgentBench~\cite{liu2024agentbench}       & \no  & \no  & \yes & \no   & \no  \\
$\tau$-bench~\cite{yao2025taubench}       & \no  & \no  & \no  & \no   & \no  \\
WixQA~\cite{cohen2025wixqa}              & \no  & \no  & \no  & \no   & \no  \\
LiveRAG~\cite{liverag2025}               & \no  & \no  & \no  & \no   & \no  \\
FinanceBench~\cite{islam2023financebench} & \no  & \no  & \no  & \no   & \no  \\
MMLongBench-Doc~\cite{ma2024mmlongbenchdoc} & \yes & \no & \no & \no   & \no  \\
TelAgentBench~\cite{lee2025telagentbench} & \no  & \yes & \no  & \yes$^\ddagger$ & \no$^\dagger$ \\
TeleQnA~\cite{maatouk2023teleqna}       & \no  & \yes & \no  & \no   & \no  \\
ToolHop~\cite{ye2025toolhop}             & \no  & \no  & \no  & \yes   & \no  \\
\midrule
\textbf{Telco-GAIA (ours)}                 & \yes & \yes & \yes & \yes & \yes \\
\bottomrule
\end{tabularx}
\caption{Comparison of key benchmark properties. Telco-GAIA is the only benchmark combining all five properties simultaneously. $^\ddagger$~partial golden-steps annotation (only a subset of tasks, or milestone checkpoints rather than reasoning steps). $^\dagger$TelAgentBench is monolingual Korean; all other prior benchmarks are monolingual English.}
\label{tab:benchmark_comparison}
\end{table*}

In summary, the principal contributions of Telco-GAIA are as follows.

\begin{enumerate}

\item \textbf{A bilingual, multi-modal, multi-hop telecom agent benchmark.}
Telco-GAIA spans three modalities (text, images, and tabular data) drawn from heterogeneous sources within one harness: a single operator's website (HTML text and on-page images), linked PDF documents (text and embedded images), a synthetic relational database (tabular data), and archival web resources (Wikipedia and ArXiv), in both Arabic and English.
No prior benchmark couples these properties (\Cref{tab:benchmark_comparison}).

\item \textbf{A reproducible, contamination-resistant environment.}
The website snapshot and customer database are frozen and served locally via Docker, so every run, whenever it is conducted, operates on an identical corpus, unlike open-internet benchmarks whose answers drift as live web content changes.

\item \textbf{Objective, deterministic scoring.}
Each of the 100~tasks has a single human-verified answer scored by normalized exact string matching, eliminating the judge-model, prompt, and scale sensitivity of LLM-as-a-Judge metrics used in RAG frameworks such as RAGAS~\cite{es2023ragas} and ARES~\cite{saadfalcon2023ares}.

\item \textbf{A relational customer database as a first-class modality.}
A synthetic SQLite database of 53~customers with twelve months of billing and usage history, queried via SQL over a REST~API, requires agents to interleave structured queries with website and PDF retrieval, a structured/unstructured combination absent from prior RAG benchmarks.

\end{enumerate}


\section{Related Work}
\label{sec:related}

Telco-GAIA sits at the intersection of several benchmark families (\Cref{tab:benchmark_comparison}).
General agent benchmarks such as GAIA~\cite{mialon2023gaia} establish exact-match, tool-using evaluation but run on the open internet, sacrificing reproducibility.
Sandboxed web environments such as WebArena~\cite{zhou2024webarena} and VisualWebArena~\cite{koh2024visualwebarena} are reproducible and general-purpose but lack a relational database and domain grounding.
Enterprise RAG benchmarks such as WixQA~\cite{cohen2025wixqa} offer a single-company corpus yet are text-only, monolingual, and without a live agent environment.
Closest in spirit, TelAgentBench~\cite{lee2025telagentbench} targets telecom agents but is Korean-only, exposes its data through abstract APIs rather than a served website, and has no image or PDF modality.
Telco-GAIA is unique in unifying all of these properties; \Cref{sec:litreview} provides a detailed description of related work. 


\section{Dataset Creation Principles}
\label{sec:specs}

This section describes the design principles and per-category specifications used to construct the 100 tasks in the Telco-GAIA Benchmark.
The dataset comprises 65~English tasks and 35~Arabic tasks, distributed across seven categories: Pricing, Miscellaneous, Images, Web Archives, PDF, PDF Visual, and Database.
The English subset was constructed first; the Arabic subset followed, built on the same principles through a combination of translation and fresh authoring.

\subsection{Common principles}

The following principles apply to every task regardless of category or language.

\paragraph{Bootstrapping by a human annotator.}
A human annotator authors a seed set of tasks, an LLM agent expands them in the same spirit, and every task, manual or synthetic, is independently validated by one or more humans.

\paragraph{Strict causal chains.}
A \emph{hop} is one retrieval-and-reasoning step, such as navigating the website to a target page, finding a fact in a PDF, querying the database, or reading a value off an image.
The reasoning process must form a single linear chain where the output of hop~$N$ is a required input for hop~$N{+}1$.
Before finalisation, every task is verified against three criteria:
(i)~every retrieved fact is a necessary input to a subsequent step (no dead ends);
(ii)~strict causal ordering holds (hop~$N{+}1$ is impossible without hop~$N$);
(iii)~no dangling or redundant facts remain.
A common failure mode is \emph{spoiling}: revealing an intermediate or final answer in an earlier step, which lets the agent skip a sub-chain of fact discovery.

\paragraph{Verification asymmetry.}
Every task must be \emph{easy to verify} (a human can confirm the answer in under 2~minutes using the \texttt{steps} field) yet \emph{hard to solve} (the agent must perform genuine search, disambiguation, and multi-tool coordination).
Difficulty comes from discovery, disambiguation, the effort to factor out potential alternative solutions, and non-obvious cross-source bridges.

\paragraph{Arabic subset.}
The Arabic subset was created by native Arabic speakers, combining human translations of existing English tasks with newly authored ones, and following the distribution of the English categories except Database.

\subsection{Category-specific principles}

\paragraph{Pricing.}
Pricing tasks retrieve exact monetary values (plan costs, device prices, add-on fees, and VAT-inclusive totals) from the operator's website. Questions imply or state the navigation path when a plan name is generic, and disambiguate Prepaid from Postpaid and Consumer from Business, whose prices differ; answers are exact numbers.

\paragraph{Miscellaneous.}
Miscellaneous tasks cover non-price reasoning: coverage checks, help-centre FAQs, device comparisons, and plan features, drawn from a deliberately diverse mix of pages. We forbid Yes/No answers. The \emph{result} of each check must drive the next hop (\eg, if a city is covered, look up its specific roaming code), keeping every step causal and the final answer a concrete value.

\paragraph{Images.}
Image tasks require the agent to locate a specific on-page visual (a banner or promotional image, identified by page and section) and extract a \emph{deterministic} fact from it: a count of people or tiers, a brand logo, or a colour. This visual fact is the first link of a 4--5-hop, multi-page chain, so a model that cannot read the image cannot proceed.

\paragraph{Web Archives.}
Web Archives tasks bridge the operator's site to external knowledge, requiring at least one fact retrieved from Wikipedia or ArXiv that is not reliably held in an LLM's parametric memory. The target entity is never named in the question (anti-spoiling): the agent must identify it from context clues and only then look it up. We prefer update-resilient targets such as infobox fields and ISO codes, and use bidirectional Wiki$\leftrightarrow$Site patterns.

\paragraph{PDF.}
PDF tasks require downloading a PDF linked from the website and extracting a specific textual value that then drives the chain (\eg, a clause number mapped to a plan ranking, then to a price). Each task draws on a PDF from a distinct product group (postpaid, prepaid, business, vacation packages) so that no document is over-exploited.

\paragraph{PDF Visual.}
PDF Visual tasks target facts that exist only in a PDF's \emph{visual layout} and not its text layer: counting coloured boxes, table rows, icons, or floor-plan elements. Every such fact is verified to be non-extractable as plain text, isolating genuine document-image understanding from mere OCR.

\paragraph{Database.}
Database tasks pair the website with a synthetic relational SQL database and demand 4--5 mostly-database hops along creative multi-source routes (DB$\to$Web$\to$DB, Web$\to$DB$\to$DB, DB$\to$Web$\to$PDF$\to$Math), with SQL ranging from multi-table joins to temporal logic. Because SQL is the agent's strongest tool, we enforce \emph{website-bridge integrity}: every cross-source bridge fact must live only on the website or in a PDF, never in a database column; otherwise the agent could answer it with SQL and skip the browser entirely (``plans-table leakage''). To keep these tasks from being trivially solvable, we further inject controlled data-quality artefacts (duplicate invoices, credit notes, internal test accounts, \texttt{NULL}-versus-zero usage) so that a careless \texttt{SELECT SUM/COUNT(*)} returns the wrong number and only an agent that inspects and filters the data succeeds; \Cref{sec:adversarial-db} details these traps and their empirically measured catch rates.


\section{Dataset Statistics}
\label{sec:stats}

\begin{figure}[htb]
\centering
\includegraphics[width=1.0\columnwidth]{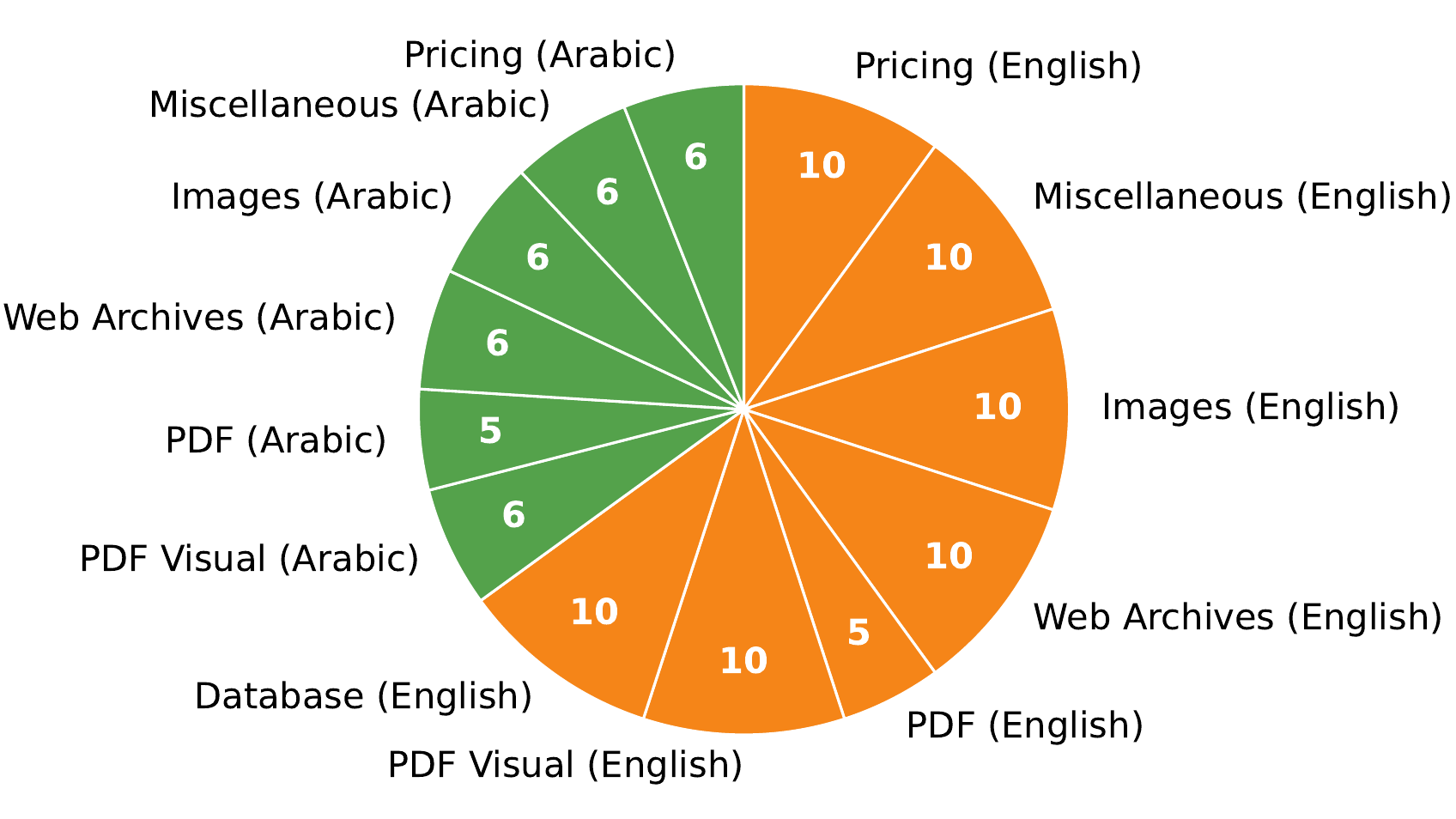}
\caption{Distribution of the tasks across categories and languages.}
\label{fig:cats_by_lang}
\end{figure}

\begin{figure}[htb]
\centering
\includegraphics[width=0.8\columnwidth]{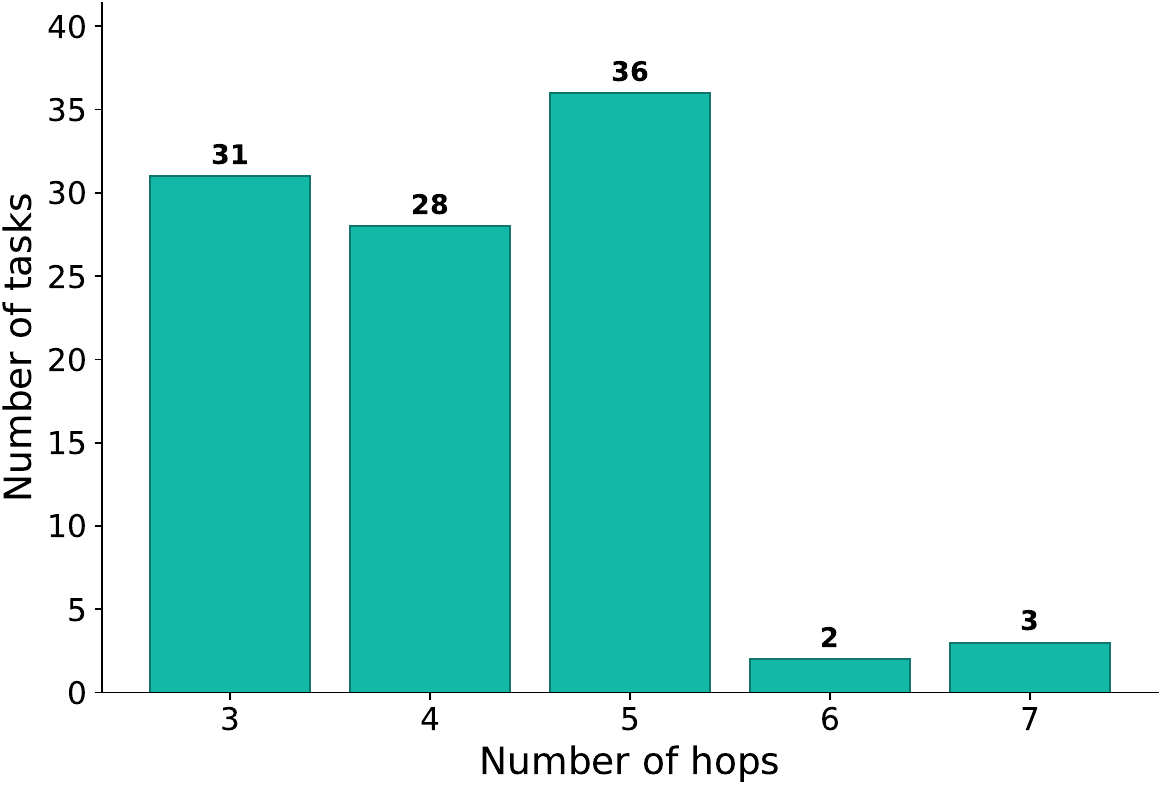}
\caption{Distribution of reasoning hops per task.}
\label{fig:hops}
\end{figure}

\Cref{fig:cats_by_lang} shows the joint distribution of the 100 tasks across the seven categories and two languages: the English subset (65 tasks) covers all seven categories, while the Arabic subset (35 tasks) mirrors six, omitting the English-only Database category. Tasks require 4.2 reasoning hops on average (\Cref{fig:hops}). Answers are 83 numeric (counts, prices, durations, codes, calculations) and 17 textual (plan names, payment methods, country names, image-derived attributes). On tooling, the local browser is universal, as every task navigates the served operator website at least once, while \texttt{pdf\_reader} (32~tasks), \texttt{internet\_browser} (16, Web Archives only), and \texttt{database\_query} (10, the entire Database category) are added as needed; 48~tasks rely on the local browser alone (\Cref{fig:tools}).


\section{Reference Agent Performance}
\label{sec:refagent}

\begin{table*}[t]
\centering
\small
\setlength{\tabcolsep}{5pt}
\begin{tabular}{lrrrrrr}
\toprule
\textbf{Model} & \textbf{Accuracy} & \textbf{Cost (\$)} & \multicolumn{2}{c}{\textbf{Duration (minutes)}} & \multicolumn{2}{c}{\textbf{Turns}} \\
\cmidrule(lr){4-5}\cmidrule(lr){6-7}
 & & & \textbf{Mean} & \textbf{Max} & \textbf{Mean} & \textbf{Max} \\
\midrule
claude-opus-4-8$^\dagger$ & 71.0\% & 176.93 & 2.0 & 37.2 & 13.9 & 100 \\
gpt-5.5            & 68.0\% & 227.71 & 34.2  & 66.7   & 17.6 & 84  \\
Kimi-K2.6$^\dagger$ & 68.0\% & 163.42 & 155.0 & 302.5  & 20.2 & 100 \\
gemini-3.5-flash   & 67.0\% & 115.29 & 2.7 & 25.1 & 19.4 & 100 \\
claude-sonnet-4-6$^\dagger$ & 67.0\% & 183.76 & 3.6 & 38.1 & 18.4 & 100 \\
DeepSeek-V4-Pro    & 48.0\% & 231.27 & 75.6  & 190.1  & 15.7 & 100 \\
gpt-5.4            & 42.0\% & 58.21  & 8.5   & 15.1   & 11.2 & 26  \\
DeepSeek-V4-Flash  & 40.0\% & 27.98  & 28.8  & 105.2  & 19.3 & 100 \\
gpt-5.2            & 38.0\% & 33.65  & 6.2   & 13.7   & 14.0 & 53  \\
gemini-3.1-flash-lite & 37.0\% & 28.86 & 2.6 & 12.3 & 39.8 & 100 \\
gpt-5.4-mini       & 15.0\% & 13.07  & 8.5   & 20.1   & 11.4 & 31  \\
gpt-5.4-nano       & 13.0\% & 3.82   & 6.4   & 12.5   & 11.4 & 34  \\
\bottomrule
\end{tabular}
\caption{Full-sweep results across twelve models on all 100~tasks (single run), sorted by accuracy. Duration and turns are per-task. $^\dagger$run under rate limits, so durations may be slightly inflated.}
\label{tab:refagent_sweep}
\end{table*}

\begin{table}[h!]
\centering
\small
\begin{tabular}{lrrr}
\toprule
\textbf{Category} & \textbf{Total} & \textbf{Accuracy} & \textbf{Std Dev} \\
\midrule
Database      & 10  & 66.0\% & 8.4\% \\
Pricing       & 16  & 62.5\% & 9.3\% \\
Web Archives  & 16  & 47.5\% & 4.4\% \\
Miscellaneous & 16  & 41.2\% & 8.8\% \\
PDF           & 10  & 26.0\% & 11.4\% \\
Images        & 16  & 25.0\% & 7.7\% \\
PDF Visual    & 16  & 3.8\%$^\ddagger$ & 4.4\% \\
\midrule
\textbf{Overall} & \textbf{100} & \textbf{38.0\%} & \textbf{2.9\%} \\
\bottomrule
\end{tabular}
\caption{Reference agent accuracy by category (GPT-5.2 backend, mean $\pm$ sample std over 5~runs). $^\ddagger$See discussion in \S\ref{sec:refagent}.}
\label{tab:refagent}
\end{table}

\begin{table}[t]
\centering
\small
\begin{tabular}{lrrr}
\toprule
\textbf{Category} & \textbf{Arabic} & \textbf{English} & \textbf{Gap} \\
\midrule
Pricing       & 69.4\% & 66.7\% & $-2.8$ \\
Miscellaneous & 45.8\% & 55.0\% & $+9.2$ \\
Images        & 23.6\% & 28.3\% & $+4.7$ \\
Web Archives  & 61.1\% & 60.0\% & $-1.1$ \\
PDF           & 48.3\% & 63.3\% & $+15.0$ \\
PDF Visual    & 18.1\% & 19.2\% & $+1.1$ \\
\midrule
\textbf{All (excl.\ DB)} & \textbf{44.3\%} & \textbf{47.4\%} & \textbf{$+3.1$} \\
\bottomrule
\end{tabular}
\caption{English vs.\ Arabic accuracy by category, averaged across the twelve models of \Cref{tab:refagent_sweep} (the Database category is English-only and excluded). Gap is English minus Arabic, in percentage points.}
\label{tab:refagent_lang}
\end{table}

\Cref{tab:refagent_sweep} reports a single-run sweep of the reference agent across twelve commercial and open models, comparing overall accuracy against API cost and per-task latency and turn counts.
Accuracy spreads cleanly from 13\% to 71\%, while cost varies by nearly two orders of magnitude (\$3.82--\$231).

The resulting cost--accuracy frontier is far from monotonic: the top models (claude-opus-4-8 71\%, gpt-5.5 68\%) sit among the more expensive, yet near-frontier accuracy is reachable far more cheaply: gemini-3.5-flash attains 67\%, within one point of gpt-5.5, at \$115: roughly half its cost and an order of magnitude lower latency.
A cluster of mid-tier models (DeepSeek-V4-Flash, gpt-5.2, gemini-3.1-flash-lite) delivers 37--40\% for under \$35, while the smallest backends (gpt-5.4-mini/nano) remain at or below 15\% despite being the cheapest to run.
Accuracy, cost, and latency thus vary along largely independent axes, so the appropriate backend depends on the deployment budget rather than on a single quality ranking.
We do not analyse absolute durations, as they depend on each model's unknown rate limits and inference provider.

The Turns columns of \Cref{tab:refagent_sweep} expose two opposite failure modes on unsolved tasks. The agent loop is capped at 100 turns (\Cref{sec:refagent_appendix}): every non-GPT backend reaches this ceiling on at least one task (maximum~=~100), with gemini-3.1-flash-lite exhausting the budget on roughly a third of its tasks (mean 39.8 turns) yet still scoring only 37\%, so these models fail by an inability to converge rather than by giving up. The GPT backends never approach the cap (maxima of 26--84 turns) and instead stop on their own; for the strongest GPT this is decisiveness (gpt-5.5 reaches 68\% in at most 84 turns), but for the weaker ones it is premature surrender, as gpt-5.2, gpt-5.4-mini, and gpt-5.4-nano halt after at most 53, 31, and 34 turns while scoring just 38\%, 15\%, and 13\%.
Either way, turn count tracks effort rather than competence: the most turn-hungry backend (gemini-3.1-flash-lite, 39.8 mean) scores only 37\%, whereas the most accurate (claude-opus-4-8, 71\%) uses just 13.9, so additional deliberation does not buy additional accuracy.
In the agentic setting, model capability outweighs sheer effort: extra turns cannot compensate for a weaker backend, so investing in a stronger model is more effective than granting a weaker one a larger budget.

To locate where this difficulty concentrates, \Cref{tab:refagent} breaks down performance by category for gpt-5.2, a medium-capacity, inexpensive backend (cf.\ \Cref{tab:refagent_sweep}) representative of a moderate cost budget, averaged over five independent runs to account for non-determinism; per-category standard deviations are propagated from per-task pass rates (\Cref{sec:variance}).

The most difficult categories are those requiring visual understanding.
Images (25.0\%) and PDF (26.0\%) both suffer from OCR-like errors: off-by-one-digit plan codes, miscounted visual elements, and confusion between marketing labels and numeric identifiers; for instance, collapsing a textual plan label into a plan-shaped token (e.g.\ ``P2+'') and emitting it in place of the numeric subscription code rather than abstaining.
A related failure is grounding on the wrong visual element: on a task asking for the colour of a specific catalogue card, every model in the sweep instead reports the colour of the page's dominant banner, defaulting to the most salient region rather than the queried target.
$^\ddagger$At 3.8\%, PDF Visual is hardest for gpt-5.2: the GPT models keep invoking the browser's unavailable PDF-download instead of the PDF reader tool, so the failure is one of tool choice rather than reading the pages; stronger vision backends do much better.

Not all category gaps reflect reasoning, however: roughly half of the Database misses by the DeepSeek backends are not wrong queries but failures to emit a parseable final answer, with a placeholder or boilerplate returned in its place. As the other eleven backends comply with the same answer-format instruction, this points to weaker instruction-following in those models rather than to the scoring.

\begin{table}[t]
\centering
\footnotesize
\setlength{\tabcolsep}{2pt}
\begin{tabular}{lrrrrr}
\toprule
\textbf{Model} & \textbf{Acc.} & \textbf{Fails} & \textbf{Early} & \textbf{Middle} & \textbf{Late} \\
\midrule
gemini-3.5-flash & 67\% & 33 & 5 (15\%) & 20 (61\%) & 8 (24\%) \\
gpt-5.2          & 38\% & 62 & 5 (8\%)  & 32 (52\%) & 25 (40\%) \\
\bottomrule
\end{tabular}
\caption{Failed tasks broken down by the earliest failing hop of the golden-steps chain: early (entry-page identification), middle (intermediate disambiguation), late (final value extraction).}
\label{tab:hop_failures}
\end{table}

The golden-steps annotations further let us localize each failure to the earliest hop at which the agent leaves the correct reasoning chain. \Cref{tab:hop_failures} reports this breakdown for two representative backends. Very few failures occur at the entry hop (8\% for gpt-5.2, 15\% for gemini-3.5-flash): both models reliably identify the right starting page. They struggle later, staying on the correct chain through intermediate disambiguation and reading off the correct final value, which are exactly the multi-hop skills the benchmark is designed to test. Notably, the stronger model does not simply make fewer errors of the same kind: the residual failures of gemini-3.5-flash skew even more heavily toward the middle hop (61\%), so its weak spot is intermediate reasoning rather than entry-point discovery or final-value extraction.

\Cref{tab:refagent_lang} reports English versus Arabic accuracy per category, averaged across the twelve models. The two languages are well balanced overall (47.4\% vs.\ 44.3\%, a 3.1-point English edge), and most categories agree within a few points, with Pricing and Web Archives even favouring Arabic slightly; the Arabic subset is therefore comparably difficult rather than systematically easier or harder. A controlled comparison excludes the English-only Database category. The clear exceptions, both favouring English, are PDF (a 15-point advantage) and Miscellaneous (9 points), the only categories with a sizeable cross-lingual gap. PDF Visual is comparable across both languages (a 1.1-point gap).

The details of the reference agent implementation can be found in \Cref{sec:refagent_appendix}.

\section{Conclusion}

We presented Telco-GAIA, a bilingual, multi-modal benchmark that evaluates tool-using agents over the public data of a real telecommunications operator. Its 100 human-verified, strictly causal multi-hop tasks draw on a served website, linked PDFs, a synthetic relational database, and external web archives, and are scored by normalized exact string matching, keeping evaluation objective, deterministic, and reproducible. Across twelve commercial and open backends the benchmark proves hard: the strongest model reaches only 71\%, a moderate-budget backend about 38\%, and the visually grounded categories lag furthest behind, exposing document and image understanding as the dominant bottleneck. We further observe that model capability, rather than expended effort, drives success, and that the English and Arabic subsets are closely matched. Beyond the resource itself, Telco-GAIA offers a reusable template for constructing reproducible, closed-domain enterprise benchmarks for the agentic era.

\section*{Limitations}

\paragraph{Compact dataset size.}
At 100~tasks Telco-GAIA is deliberately compact: the small size limits statistical power and fine-grained per-category comparison, but each task is a densely annotated, shortcut-free multi-hop chain (mean 4.2, up to 7~hops) with a human-verified solution path, echoing other depth-over-scale benchmarks (TheAgentCompany~\cite{xu2025theagentcompany}, 175~tasks; $\tau$-Knowledge~\cite{shi2026tauknowledge}, 97~tasks; WorkArena~\cite{drouin2024workarena}, 33~templates).
The same logic applies to language: rather than scaling the Arabic subset by machine translation, which injects noise even under careful control~\cite{alrashed2024finewebeduar}, every Arabic task is human-translated or natively authored, leaving language coverage uneven (65~English vs.\ 35~Arabic tasks, with the Database category English-only).
Evaluation stays cheap regardless: a full run costs from a few dollars to under \$300 per model (\Cref{tab:refagent_sweep}).

\paragraph{Single operator and domain.}
Built from the corpus of a single operator in one market, Telco-GAIA is realistic and internally consistent but may not transfer to other operators, industries, or regulatory contexts; we thus present it as a difficulty probe and a template for closed enterprise benchmarks rather than a measure of general enterprise competence.

\paragraph{Synthetic customer database.}
The relational customer database is fully synthetic, a deliberate choice that avoids exposing personal or commercially sensitive data and keeps the benchmark freely shareable.
Consequently, although we deliberately inject controlled data-quality artefacts (\Cref{sec:adversarial-db}), the Database category does not capture the full scale or organic noise of production telecom data.

\publiconly{
\section*{Acknowledgments}

The research was funded by Sponsored Research Agreement ORA \#6323 between King Abdullah University of Science and Technology (KAUST) and Saudi Telecommunications Company (stc).
}

\FloatBarrier


\bibliography{custom}

\appendix

\section{Benchmark Harness}
\label{sec:harness_appendix}

\begin{figure}[htb]
\centering
\includegraphics[width=1.0\columnwidth]{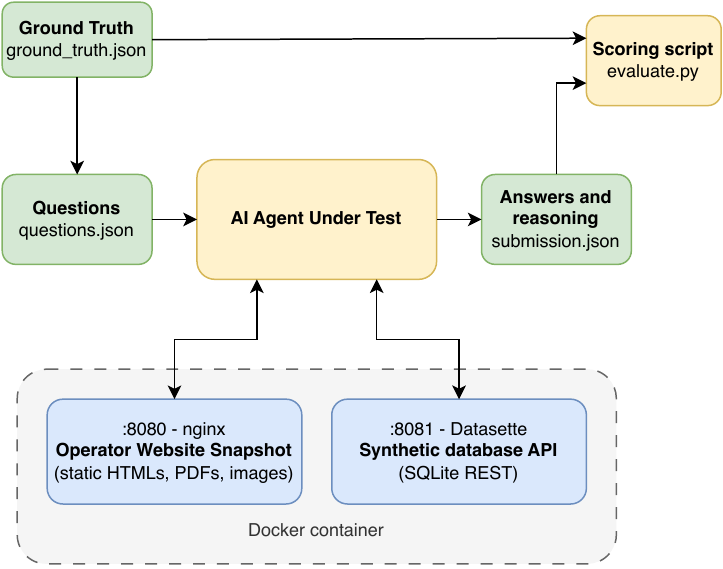}
\caption{Benchmark architecture. The agent reads task inputs and queries the Docker container; the evaluator scores the submission against ground truth.}
\label{fig:architecture}
\end{figure}

The benchmark is packaged as a self-contained Docker environment that exposes the local website and the SQL database to the agent via local HTTP endpoints, while Wikipedia and ArXiv are accessed directly on the internet. Answers are evaluated by \textbf{exact string matching} after minimal normalization, with no LLM-as-a-Judge at any stage. \Cref{fig:architecture} illustrates the evaluation pipeline.

\subsection{Docker Container}

The container (\texttt{bench\_compose}) bundles two services managed by \texttt{supervisord}:

\begin{itemize}
\item \textbf{Port 8080 --- nginx}: Serves the static website snapshot (HTML pages, images, PDFs) from the February~2026 crawl. Both English and Arabic site trees are available.
\item \textbf{Port 8081 --- Datasette}: Exposes the synthetic SQLite customer database as a read-only REST~API with JSON output, supporting read-only SQL queries.
\end{itemize}

\subsection{Agent Inputs}

The agent under test receives two files (never the ground truth):

\begin{itemize}
\item \texttt{questions.json} --- a JSON array of 100 tasks, each containing only \texttt{task\_id} and \texttt{question}.
\item \texttt{environment.md} --- a Markdown document describing the available tools and endpoints.
\end{itemize}

A \emph{task} is the unit of evaluation: its \texttt{question} field is the natural-language prompt shown to the agent, while its full serialization across all fields below is the ground-truth \emph{record}.
Each full ground-truth record carries the fields \texttt{task\_id}, \texttt{category}, \texttt{language}, \texttt{question}, \texttt{final\_answer}, \texttt{answer\_type}, \texttt{steps}, \texttt{num\_steps}, \texttt{tools}, and \texttt{num\_tools}; only \texttt{task\_id} and \texttt{question} are exposed to the agent.
The harness additionally restricts the reference agent to each task's permitted tool subset (its \texttt{tools} field; see \Cref{sec:refagent_appendix}). This is task metadata used only to gate which tools are exposed; it never reveals the answer.

\subsection{Submission Format}

The agent must produce a \texttt{submission.json} in GAIA-compatible format:

\begin{verbatim}
[
  {
    "task_id": "task-001",
    "model_answer": "50",
    "reasoning_trace": "I navigated to..."
  }
]
\end{verbatim}

Each entry must include \texttt{task\_id} and \texttt{model\_answer}; the \texttt{reasoning\_trace} field is optional but recommended for diagnostic purposes.

\subsection{Scoring}

Evaluation is performed by \texttt{evaluate.py}, a pure-Python script with no external dependencies:

\begin{verbatim}
python evaluate.py \
    --submission path/to/submission.json \
    --ground-truth ground_truth.json \
    --output results.json
\end{verbatim}

The scorer applies GAIA-convention normalization (lowercasing, whitespace stripping, removal of currency symbols and commas) and performs exact string matching.
Each task receives a binary score (1~if correct, 0~otherwise).
The output \texttt{results.json} contains the overall accuracy, per-task scores, and a category breakdown reporting accuracy for each of the seven categories independently.
Missing submissions are scored as~0; extra task IDs trigger a warning but do not affect scoring.

\paragraph{Exact-match error audit.}
A manual audit of the reference agent's answers found that ${\sim}3\%$ land close to the ground truth but violate the answer format the question demands and are counted incorrect by design; we encountered no correct answer penalized for legitimate phrasing variance. Question wording was iterated during dataset creation until the expected answer format was unambiguous. Users preferring graded scoring can unofficially layer an LLM- or agent-based judge~\cite{zhuge2025agentasjudge} on top of the released ground truth; the official metric remains normalized exact match.

\publiconly{
\subsection{Local Services in Action}

Figures~\ref{fig:local_website} and~\ref{fig:local_db} show the two containerized services as they appear to the agent at runtime.

\begin{figure}[h!]
\centering
\includegraphics[width=1.0\columnwidth]{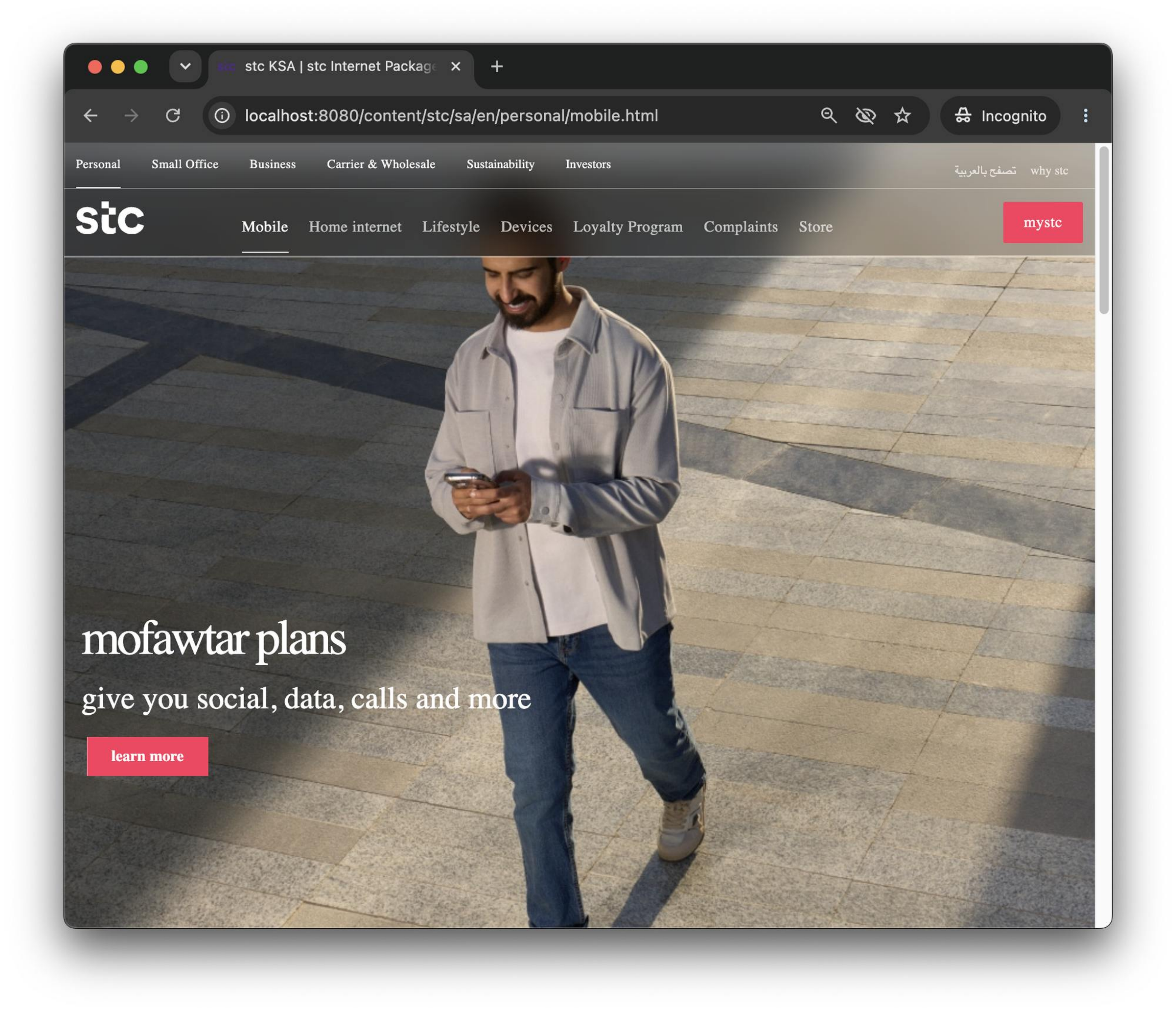}
\caption{The operator website served locally via nginx on port~8080. The agent navigates this static snapshot exactly as it would the live site.}
\label{fig:local_website}
\end{figure}

\begin{figure}[h!]
\centering
\includegraphics[width=1.0\columnwidth]{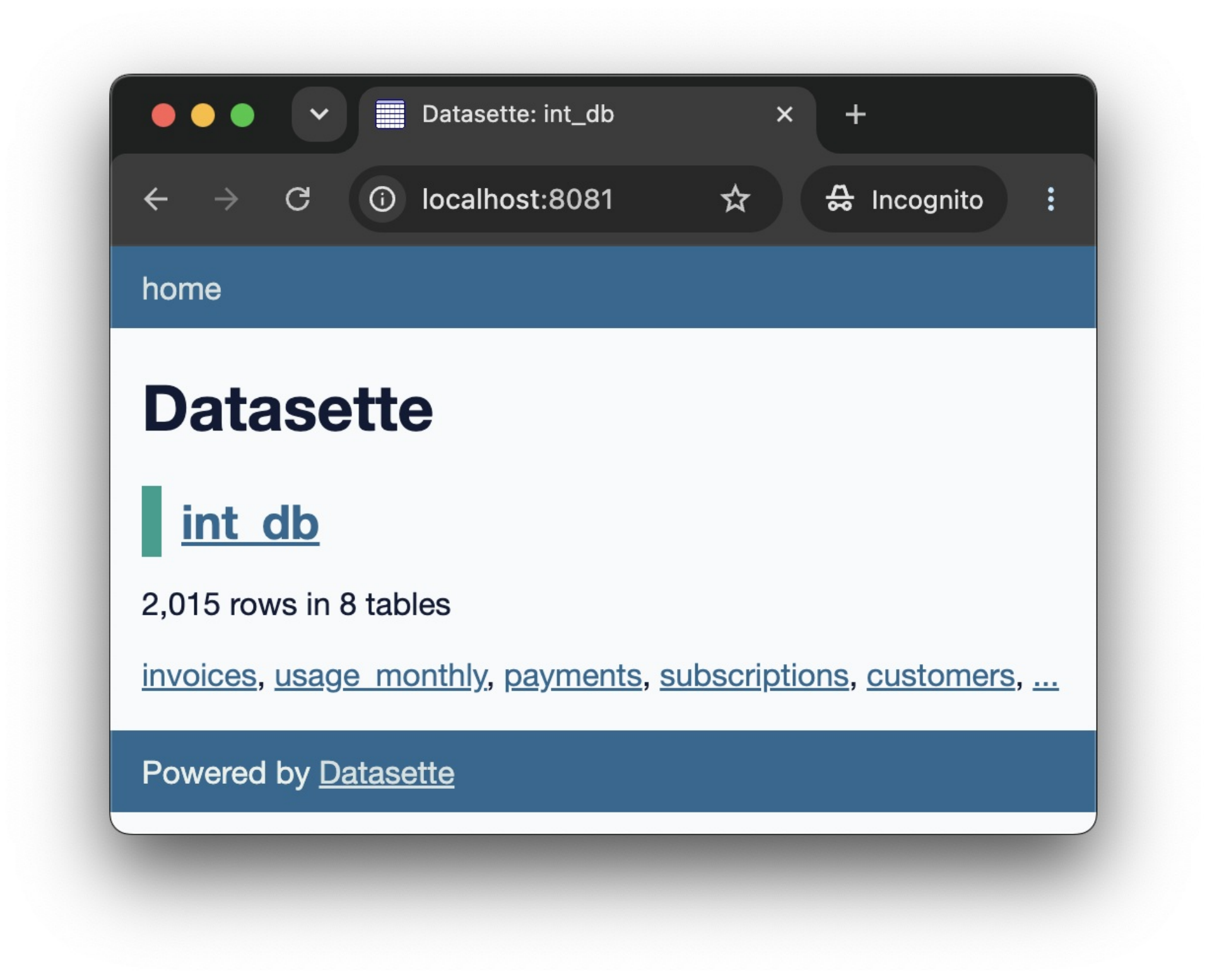}
\caption{The Datasette interface on port~8081 exposing the synthetic SQLite database (\texttt{int\_db}) with 2{,}015 rows across 8~tables.}
\label{fig:local_db}
\end{figure}
}

\section{Adversarial Database Design}
\label{sec:adversarial-db}

A clean database is trivially solvable: once an agent discovers the schema, a single \texttt{SELECT} returns the answer. To make database reasoning genuinely hard, we deliberately inject controlled data-quality artefacts (the kind that accumulate in any production billing system) so that a naive \texttt{SELECT SUM/COUNT(*)} returns the \emph{wrong} number while the gold answer is unchanged. Only an agent that inspects the data, detects the anomaly, and applies the right deduplication or filter can recover the correct value. The artefacts span ten families, including duplicate (double-fired) invoices, credit-note rows with negative amounts, internal test accounts, and \texttt{NULL}-versus-zero usage; every database task triggers at least one.

\paragraph{Wording is the control knob.}
The difficulty of a trap is governed not by the data alone but by the question's wording. Over-explicit phrasing (``excluding duplicates and credit notes'') hands the agent the exact filter and neutralises the trap, while over-implicit phrasing risks ambiguity for human reviewers. The sweet spot is a \emph{subtle} qualifier (``the legitimate invoiced amount'', ``all real active subscribers'', ``since the subscription started'') that a careful human acts on by inspecting the data, but that an agent cannot mechanically compile into a SQL clause. The wording must keep the gold answer the only defensible reading while leaving the trap live.

\paragraph{Not all traps are equal.}
A five-fold evaluation of the reference agent lets us grade traps by empirical catch rate, which spans 100\% down to 0\%. The decisive factor is whether a trap poisons the agent's \emph{inevitable} query path: an internal test account that out-ranks the true customer (or a subtle qualifier that demands data inspection) is caught on essentially every run, whereas a trap that merely hopes the agent forgets a status filter is probabilistic (40--60\%), and explicit wording that spells out the filter neutralises it entirely.

\section{Reference Agent Implementation}
\label{sec:refagent_appendix}

The reference agent is a tool-calling LLM loop built on LiteLLM; the implementation is open-source.\publiconly{\footnote{\url{https://github.com/kaust-generative-ai/telco_gaia_ref_agent}}}
The scaffold is deliberately simple and hand-designed; richer agentic scaffolds can also be optimized automatically, \eg\ as computational graphs~\cite{zhuge2024gptswarm} or via semantic backpropagation~\cite{wang2024semanticbp}, and Telco-GAIA's deterministic scores provide a stable objective for such methods.
At each iteration the model receives a task-specific system prompt, selects a tool, observes the result, and decides whether to call another tool or emit a final answer.
Four tools are available, but for each task the reference agent is restricted to the subset listed in that task's \texttt{tools} field (\eg, a Web Archives task exposes \texttt{local\_browser} and \texttt{internet\_browser} only). This gating prevents shortcut solutions that bypass the intended reasoning chain, such as reading a plan price directly from the database when the task requires retrieving it from the website. The model receives schemas only for these permitted tools. The four tools are:

\begin{itemize}
\item \texttt{local\_browser} --- a Playwright-based headless browser restricted to the served operator website (\texttt{localhost:8080}).
\item \texttt{internet\_browser} --- a second Playwright instance restricted to \texttt{*.wikipedia.org} and \texttt{*.arxiv.org} for Web Archives tasks.
\item \texttt{pdf\_reader} --- downloads PDFs and extracts content via \texttt{pdfplumber} (text layer) with a \texttt{PyMuPDF} vision fallback for visual PDFs.
\item \texttt{database\_query} --- executes read-only SQL against the Datasette API on \texttt{localhost:8081}.
\end{itemize}

\begin{figure}[htb]
\centering
\includegraphics[width=1.0\columnwidth]{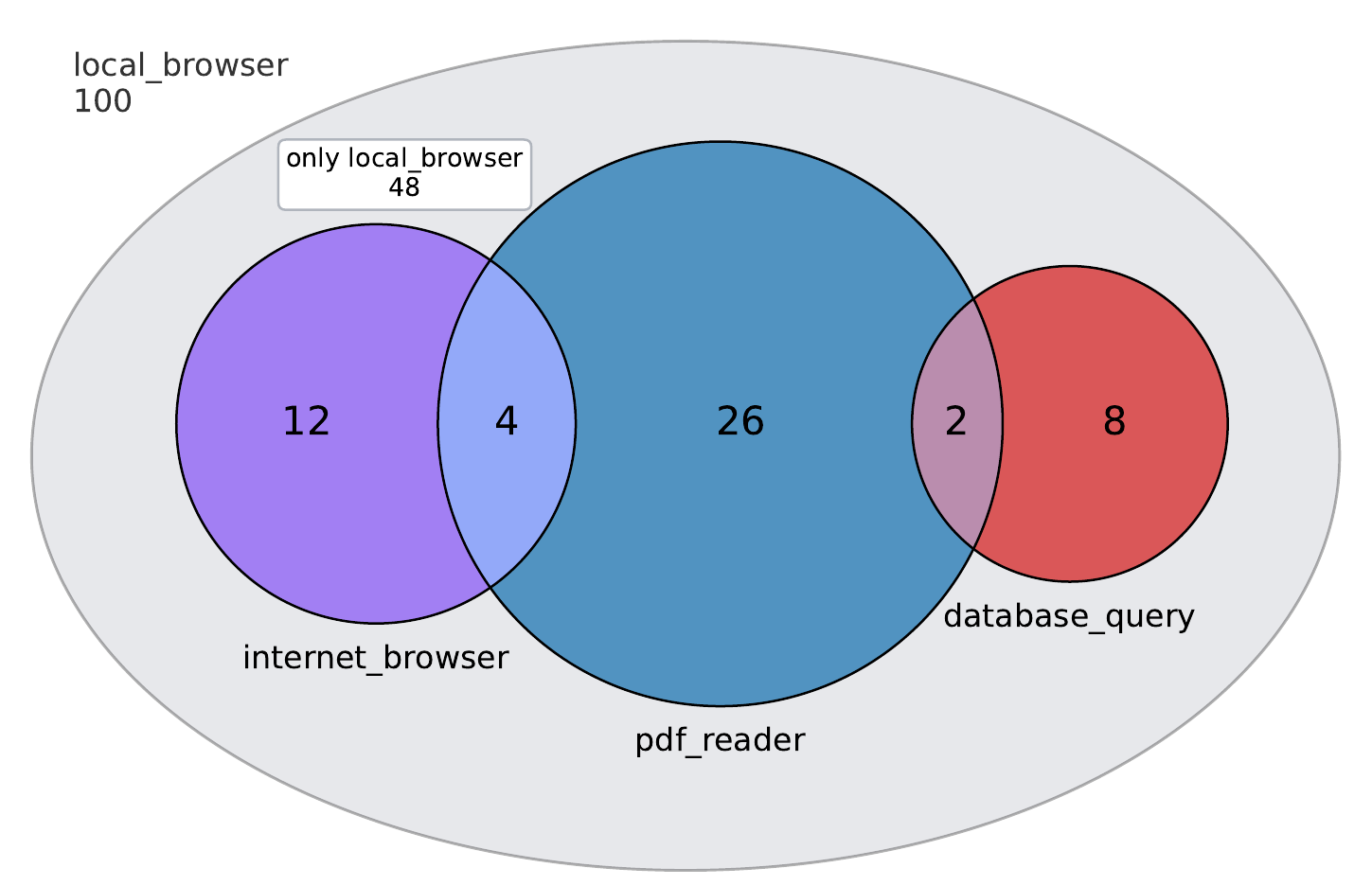}
\caption{Tool usage across the 100 tasks. The grey \texttt{local\_browser} ellipse is a superset (used by every task).}
\label{fig:tools}
\end{figure}

\Cref{fig:tools} shows how these four tools are distributed across the 100 tasks.

The agent outputs results in the GAIA-compatible submission format described in \Cref{sec:harness_appendix}.

\paragraph{Evaluated backends.}
The full sweep (\Cref{tab:refagent_sweep}) covers twelve commercial and open backends: gpt-5.5, gpt-5.4, gpt-5.2, gpt-5.4-mini, and gpt-5.4-nano~\cite{openai2025gpt5}; gemini-3.5-flash and gemini-3.1-flash-lite~\cite{google2025gemini3}; claude-opus-4-8~\cite{anthropic2026claudeopus48} and claude-sonnet-4-6~\cite{anthropic2026claudesonnet46}; Kimi-K2.6~\cite{kimiteam2025kimik2}; and DeepSeek-V4-Pro and DeepSeek-V4-Flash~\cite{deepseek2026v4}.

\subsection{Score variance estimation}
\label{sec:variance}

For each task~$i$, we estimate its solve probability $\hat{p}_i$ as its pass rate across the $N{=}5$ runs, yielding sample variance $\hat{\sigma}_i^2 = \hat{p}_i(1-\hat{p}_i)\,N/(N{-}1)$.
The category mean accuracy is $\bar{p} = \frac{1}{T}\sum_i \hat{p}_i$, and since task outcomes are independent across runs, the standard deviation of the category score is $\sigma = \frac{1}{T}\sqrt{\sum_i \hat{\sigma}_i^2}$, where $T$~is the number of tasks in the category.
This task-level variance propagation avoids under-estimating variability when different tasks fluctuate but their totals coincidentally cancel.

\subsection{Parametric-knowledge baseline}
\label{sec:pk_mode}

To quantify how much of the benchmark is solvable from parametric memory alone (the operator's website is public and may appear in pretraining corpora), we re-ran the reference agent with all tools and data access disabled, forcing each backend to answer directly. \Cref{tab:pk_mode} contrasts the two regimes: scores collapse to 1--18\%. Most of the residual comes from the Web Archives category, whose Wikipedia-derived facts are well represented in pretraining data. This leakage also decays over time: as the operator adjusts plans and prices, stale parametric knowledge diverges from the frozen snapshot, further hardening the benchmark.

\begin{table}[h!]
\centering
\small
\begin{tabular}{lrr}
\toprule
\textbf{Model} & \textbf{With tools} & \textbf{No tools} \\
\midrule
claude-opus-4-8       & 71.0\% & 8.0\% \\
gpt-5.5               & 68.0\% & 18.0\% \\
gemini-3.5-flash      & 67.0\% & 15.0\% \\
claude-sonnet-4-6     & 67.0\% & 8.0\% \\
gpt-5.4               & 42.0\% & 6.0\% \\
gpt-5.2               & 38.0\% & 8.0\% \\
gemini-3.1-flash-lite & 37.0\% & 7.0\% \\
gpt-5.4-mini          & 15.0\% & 4.0\% \\
gpt-5.4-nano          & 13.0\% & 1.0\% \\
\bottomrule
\end{tabular}
\caption{Reference-agent accuracy with full tool access versus with all tools and data access disabled (parametric knowledge only).}
\label{tab:pk_mode}
\end{table}


\section{Literature Review}
\label{sec:litreview}

Evaluation benchmarks for AI agents and retrieval-augmented generation (RAG) systems have proliferated rapidly since 2023.
We organize the relevant prior work into five clusters, each illuminating a different dimension of the design space.


\subsection{General AI Agent Benchmarks}

\textbf{GAIA}~\cite{mialon2023gaia} (Meta/HuggingFace, ICLR~2024) is the direct inspiration for Telco-GAIA.
It comprises 466 real-world questions organized into three difficulty levels, requiring general-purpose tool use including open internet browsing, file parsing (PDF, spreadsheet, audio), and multi-step reasoning.
Questions are hand-crafted and non-synthetic; answers are exact strings verified by human annotators.
Human performance is 92\%, while GPT-4 equipped with plugins achieved only $\sim$15\% at launch, with top agents since reaching $\sim$75\%.
The key distinction from Telco-GAIA is that GAIA is fully open-domain and relies on the live internet, making it non-reproducible across time and inapplicable to closed enterprise corpora.

\textbf{AssistantBench}~\cite{yoran2024assistantbench} (2024) contains 214 realistic web tasks (\eg, monitoring real-estate listings, discovering local businesses) evaluated on the open internet.
Like GAIA, it uses automatic evaluation and reveals that no model exceeded 25 accuracy points at publication time.
It has no domain focus, no database modality, and no sandboxed environment.

\textbf{BrowseComp}~\cite{browsecomp2025} (OpenAI, 2025) consists of 1,266 questions requiring persistent internet navigation to locate hard-to-find, multi-hop facts.
It is an open-web benchmark measuring search persistence and creativity rather than domain-specific knowledge.


\subsection{Sandboxed Web Agent Environments}

\textbf{WebArena}~\cite{zhou2024webarena} (ICLR~2024) is the closest architectural ancestor of Telco-GAIA.
It constructs fully sandboxed, Docker-hosted realistic websites across four domains: e-commerce, social forum, code hosting, and content management.
Agents interact via browser automation (UI clicks and form submission); 812 tasks are evaluated.
The best GPT-4-based agent achieved only 14.4\% success rate versus 78\% for humans.
Unlike Telco-GAIA, WebArena requires browser-level UI interaction rather than HTTP/API tool use, has no dedicated relational database modality, no PDF or image retrieval tasks, and is general-purpose rather than domain-specific.

\textbf{WorkArena}~\cite{drouin2024workarena} (ServiceNow/ICML~2024) evaluates LLM agents on 33 enterprise knowledge-work tasks inside a live ServiceNow instance (\eg, filling forms, ordering from service catalogs, reading dashboards).
It introduces BrowserGym, an environment with multimodal observations.
GPT-4 achieves 42.7\%.
Its task instances are programmatically generated from parameterized templates rather than hand-authored.
WorkArena is limited to UI interaction inside a single SaaS platform and has no external retrieval corpus or structured database modality.

\textbf{AgentBench}~\cite{liu2024agentbench} (ICLR~2024) is a comprehensive multi-environment suite covering 8 agent tasks: operating system shell, database (SQL), knowledge graph, card games, lateral thinking puzzles, web shopping, web browsing, and household navigation.
It evaluates 29 LLMs and finds a large gap between commercial and open-source models.
While AgentBench includes a SQL environment, it does not combine structured and unstructured retrieval in a single coherent domain scenario.

\textbf{VisualWebArena}~\cite{koh2024visualwebarena} (ACL~2024) extends WebArena with 910 visually grounded web tasks requiring multimodal agents to interpret images, screenshots, and visual page layouts.
It introduces execution-based visual evaluation metrics but retains WebArena's general-purpose multi-domain design and lacks PDF, database, or domain-specific retrieval.

\textbf{TheAgentCompany}~\cite{xu2025theagentcompany} (NeurIPS~2025) creates a self-contained simulated software company with internal websites, code repositories, and communication tools.
It evaluates agents on 175 consequential professional tasks (writing code, browsing internal docs, communicating with simulated coworkers); the best agent completes only 30\% autonomously.
Unlike Telco-GAIA, it targets general office work rather than domain-specific QA and has no structured database modality or PDF retrieval.

\textbf{$\tau$-bench}~\cite{yao2025taubench} (Sierra/Princeton, ICLR~2025) evaluates tool-calling agents on realistic customer-service dialogues in retail and airline domains, with a telecom domain added in 2025 ($\tau^2$-bench~\cite{barres2025tau2bench}).
GPT-4o achieves only $\sim$35\% pass@1 in the airline domain ($\sim$61\% in retail) and exhibits poor consistency (pass@8 $<$ 25\% in retail); the harder telecom domain introduced in $\tau^2$-bench drops even GPT-4.1 to $\sim$34\%.
Its domain databases and user tasks are synthetically generated with GPT-4.
Unlike Telco-GAIA, $\tau$-bench focuses on conversational tool use rather than multi-source retrieval, has no website or PDF modality, and tests single-turn tool calls rather than multi-hop reasoning chains.

\textbf{OSWorld}~\cite{xie2024osworld} (NeurIPS~2024) evaluates multimodal agents on 369 real-world computer tasks spanning Ubuntu, Windows, and macOS using virtual machines.
It targets full desktop control rather than retrieval-focused evaluation.


\subsection{Enterprise RAG Benchmarks}

\textbf{WixQA}~\cite{cohen2025wixqa} (Wix.com, May~2025) is the most structurally similar RAG benchmark to Telco-GAIA.
It provides three datasets grounded in a snapshot of the Wix Help Center knowledge base (6,220 articles): 200 expert-written pairs, 200 expert-validated simulated pairs, and 6,222 LLM-generated synthetic pairs.
The KB snapshot is released alongside the dataset under MIT licence.
WixQA differs from Telco-GAIA in that it is a pure text RAG benchmark: there is no live agent environment, no relational database, no image or PDF modality, and no bilingual (Arabic/English) content.

\textbf{RAGBench}~\cite{friel2024ragbench} (2024) provides 100k evaluation examples across five industry-specific corpora (user manuals, biomedical texts, finance, legal, conversational AI).
It introduces the TRACe evaluation framework with explainability metrics.
RAGBench is large-scale but is a static corpus dump without an agent environment or structured data sources.

\textbf{CRAG}~\cite{yang2024crag} (2024) is a comprehensive RAG benchmark covering multiple domains and task types, introduced as a KDD Cup competition.
It assesses retrieval quality, factual accuracy, and reasoning, but targets open-domain web retrieval rather than a closed enterprise corpus.

\textbf{ARES}~\cite{saadfalcon2023ares} (2023) is an automated evaluation framework for RAG systems that uses synthetic data generation and fine-tuned classifiers as judges.
It is a meta-framework rather than a fixed-corpus benchmark.

\textbf{LiveRAG}~\cite{liverag2025} (TII/SIGIR~2025) is a competition-derived benchmark of 895 synthetic questions over a fixed 10-billion-token web corpus (FineWeb-10BT), with difficulty scores calibrated from 70 competing teams.
It enforces a fixed retrieval corpus and fixed LLM (Falcon3-10B), isolating RAG pipeline quality.
Unlike Telco-GAIA, LiveRAG is open-domain, text-only, and evaluates retrieval pipeline components rather than multi-modal agent reasoning.


\subsection{Multi-Hop Retrieval Benchmarks}

\textbf{HotpotQA}~\cite{yang2018hotpotqa} (2018) contains 112,779 Wikipedia-based questions requiring 2-hop reasoning with supporting sentence annotations.
It is the canonical multi-hop QA baseline but has no agent environment and no closed or domain-specific corpus.

\textbf{MuSiQue}~\cite{trivedi2022musique} (2022) systematically composes single-hop questions into 25k 2--4-hop questions to prevent reasoning shortcut exploitation.
It exhibits a 3$\times$ larger human-machine gap than HotpotQA.
Like HotpotQA, it is Wikipedia-only with no agent environment.

\textbf{FRAMES}~\cite{krishna2024frames} (Google DeepMind, NAACL~2025) contains 824 challenging multi-hop questions each requiring information from 2--15 Wikipedia articles.
It provides a unified evaluation of factual accuracy, retrieval, and reasoning.
Multi-step retrieval improves accuracy from 47\% (single step) to 66\%.

\textbf{FanOutQA}~\cite{zhu2024fanoutqa} (2024) comprises 1,034 multi-document Wikipedia questions with human-written decompositions.
It evaluates closed-book, open-book, and evidence-provided settings.

\textbf{MultiHop-RAG}~\cite{tang2024multihoprag} (2024) focuses on multi-hop RAG evaluation using English news articles, evaluating both the retriever and the reader components.

\textbf{BRIGHT}~\cite{su2024bright} (ICLR~2025) contains 1,385 reasoning-intensive retrieval queries drawn from StackExchange, LeetCode, and math competitions.
State-of-the-art retrievers score only 18 nDCG@10, demonstrating the difficulty of reasoning-intensive retrieval.

\textbf{MINTQA}~\cite{mintqa2024} (2024) evaluates LLMs on 28k multi-hop questions specifically targeting \emph{new} knowledge (post-training-cutoff facts) and \emph{tail} knowledge (rare entities).
It tests four QA strategies including iterative decomposition with RAG.
Unlike Telco-GAIA, MINTQA is Wikipedia-based with no agent environment, no tool use, and no domain-specific corpus.

\textbf{ToolHop}~\cite{ye2025toolhop} (ByteDance, ACL~2025) is the benchmark most directly comparable to Telco-GAIA's multi-hop design philosophy.
It comprises 995 user queries paired with 3,912 tools, where each query requires sequential tool invocations with meaningful interdependencies between hops, in which the output of one tool call serves as input to the next.
GPT-4o, the best-performing model, achieves only 49\% accuracy.
ToolHop validates that multi-hop tool chaining remains a frontier challenge.
However, it uses synthetic tools rather than real-world retrieval sources, has no domain grounding, no website navigation, and no structured database or PDF modality.


\subsection{Domain-Specific and Structured Data Benchmarks}

\textbf{FinanceBench}~\cite{islam2023financebench} (Patronus~AI, 2023) contains 10,231 questions grounded in SEC~filings (PDFs) about publicly traded companies.
GPT-4 with a retrieval system incorrectly answered or refused 81\% of questions.
FinanceBench is a single-domain, text-based, closed-corpus benchmark with no agent environment or database modality; evidence is the extracted page text of filings (not their visual layout), and a substantial share of its questions are programmatically generated from financial metrics rather than hand-written.

\textbf{SEC-QA}~\cite{wang2024secqa} (2024) extends the financial QA paradigm to multi-document questions spanning multiple long-context filings.
It introduces a continuous dataset refresh capability to prevent data contamination.

\textbf{MMLongBench-Doc}~\cite{ma2024mmlongbenchdoc} (2024) provides 1,082 expert-annotated questions across 135 long PDF documents (avg.\ 47.5 pages) requiring reasoning over text, tables, figures, and layout structure.
33\% of questions require cross-page evidence; even GPT-4o achieves only 44.9\% F1.
MMLongBench-Doc is the closest benchmark to Telco-GAIA's PDF Visual category, but it has no agent environment, no website navigation, no database, and is English-only.

\textbf{TelAgentBench}~\cite{lee2025telagentbench} (SK~Telecom, EMNLP~2025 Industry) is the telecom-domain benchmark most closely related to Telco-GAIA.
It is a Korean-language synthetic benchmark evaluating LLM agents on five capabilities: Reasoning, Planning, Action (tool use), RAG, and Instruction Following, with over 1,700 instances.
It provides a sandbox environment of 23 telecom business-support-system (BSS) APIs backed by a simulated database, all grounded in SK~Telecom's own service data. Unlike Telco-GAIA, however, it exposes this environment through abstract function-call APIs rather than a scraped served website, does not expose a relational database that the agent queries directly, has no image or PDF modality, and is Korean-only.

\textbf{TeleQnA}~\cite{maatouk2023teleqna} (2023) is the foundational telecom knowledge benchmark, comprising 10,000 multiple-choice questions across five categories: lexicon, research overview, research publications, standards overview, and standards specifications.
GPT-4 achieved 64\% on the hardest standards category.
Its question-answer pairs are produced by an automated LLM-based generation pipeline with human quality review.
TeleQnA tests static domain knowledge rather than agent capabilities; it has no tool use, no retrieval environment, and no multi-hop reasoning.

\textbf{TeleTables}~\cite{teletables2025} (Jan~2026) provides 500 human-verified question-answer pairs for evaluating LLMs on table interpretation in 3GPP technical specifications.
It targets a specific modality (structured tables in standards documents) and demonstrates that smaller models ($<$10B parameters) fail on complex telecom table reasoning.
Unlike Telco-GAIA, it is a static QA benchmark with no agent environment.

\textbf{TelBench}~\cite{telbench2024} (EMNLP~2024 Industry) evaluates telco-specific LLMs on customer service expertise across 34 models, but is a text-only evaluation without an agent environment.

\textbf{TeleEval-CS}~\cite{teleeval2024} (2024) provides 8.1k customer-service examples across 15 subtasks and three call stages, also without an agent environment.


\end{document}